\documentclass[10pt,twocolumn,letterpaper]{article}

\usepackage{iccv}
\usepackage{times}
\usepackage{epsfig}
\usepackage{graphicx}
\usepackage{amsmath}
\usepackage{amssymb}
\usepackage{caption}
\usepackage{subcaption}
\usepackage[sort, numbers, compress, square]{natbib}
\usepackage{amsthm}
\usepackage{pifont}
\usepackage{booktabs}
\usepackage{xcolor}
\usepackage[accsupp]{axessibility}  

\newcommand{\cmark}{\ding{51}}%
\newcommand{\xmark}{\ding{55}}%

\usepackage[pagebackref=true,breaklinks=true,letterpaper=true,colorlinks,bookmarks=false]{hyperref}

\iccvfinalcopy 

\def\iccvPaperID{11645} 

\ificcvfinal\fi

\begin{document}

\title{Generalizable Neural Fields as Partially Observed Neural Processes}

\author{Jeffrey Gu\\
ICME\\
Stanford, CA\\
{\tt\small jeffgu@stanford.edu}
\and
Kuan-Chieh Wang\\
Department of Computer Science\\
Stanford, CA\\
{\tt\small wangkua1@stanford.edu}
\and
Serena Yeung\\
Department of Computer Science\\
Stanford, CA\\
{\tt\small syyeung@stanford.edu}
}

\maketitle
\ificcvfinal\thispagestyle{empty}\fi

\begin{abstract}
   Neural fields, which represent signals as a function parameterized by a neural network, are a promising alternative to traditional discrete vector or grid-based representations. Compared to discrete representations, neural representations both scale well with increasing resolution, are continuous, and can be many-times differentiable. However, given a dataset of signals that we would like to represent, having to optimize a separate neural field for each signal is inefficient, and cannot capitalize on shared information or structures among signals. Existing generalization methods view this as a meta-learning problem and employ gradient-based meta-learning to learn an initialization which is then fine-tuned with test-time optimization, or learn hypernetworks to produce the weights of a neural field. We instead propose a new paradigm that views the large-scale training of neural representations as a part of a partially-observed neural process framework, and leverage neural process algorithms to solve this task. We demonstrate that this approach outperforms both state-of-the-art gradient-based meta-learning approaches and hypernetwork approaches. 
\end{abstract}

\section{Introduction}

Neural fields, also known as neural implicit representations or coordinate-based neural networks, have shown great promise as an alternative to traditional discrete representations. Neural fields consider signals and other objects as functions (\textit{fields}) mapping coordinates to values, and approximate these functions with neural networks. 
They have been used to represent a myriad of signals including 3D objects/scenes \cite{mildenhall2021nerf,sitzmann2019scene}, CT scans \cite{tancik2021learned}, and 3D human motion \cite{wang2023nemo}.
This approach has several advantages over discrete representations, including being continuous and having non-zero derivatives \cite{sitzmann2020implicit} (depending on the choice of neural network architecture), and scale much more efficiently than traditional grid-based representations with increasing resolution \cite{park2019deepsdf, mescheder2019occupancy, chen2019learning, genova2019learning, atzmon2020sal, peng2020convolutional, sitzmann2019scene}.

\begin{figure*}[h]
    \centering
    \begin{subfigure}{.5\textwidth}
        \centering
        \includegraphics[width=\textwidth]{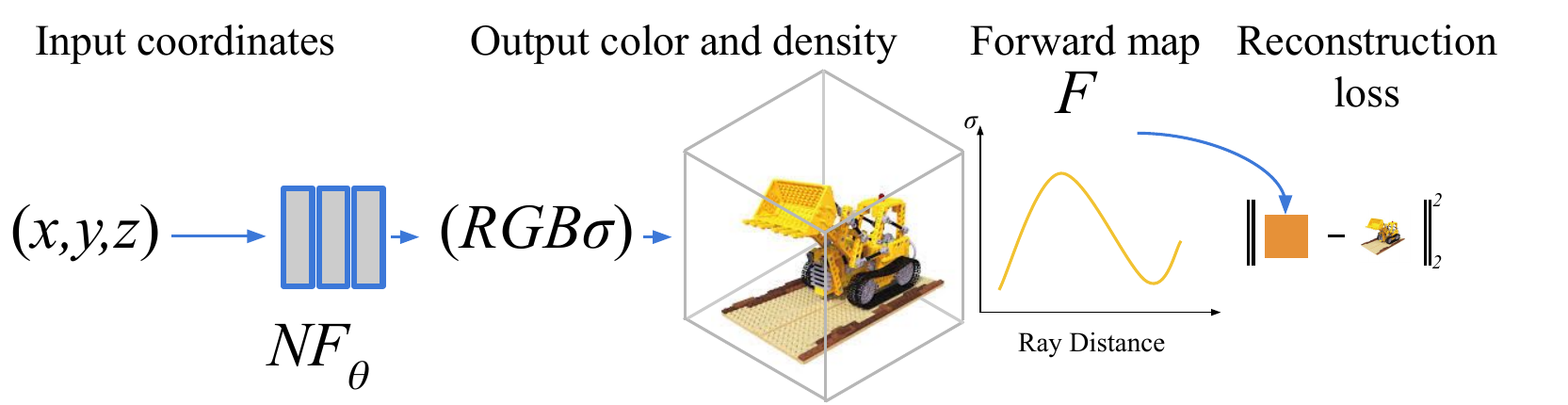}
        \caption{Typical neural field training framework}
        \label{fig:typical-nf-training}
    \end{subfigure}%
    \begin{subfigure}{.5\textwidth}
        \centering
        \includegraphics[width=\textwidth]{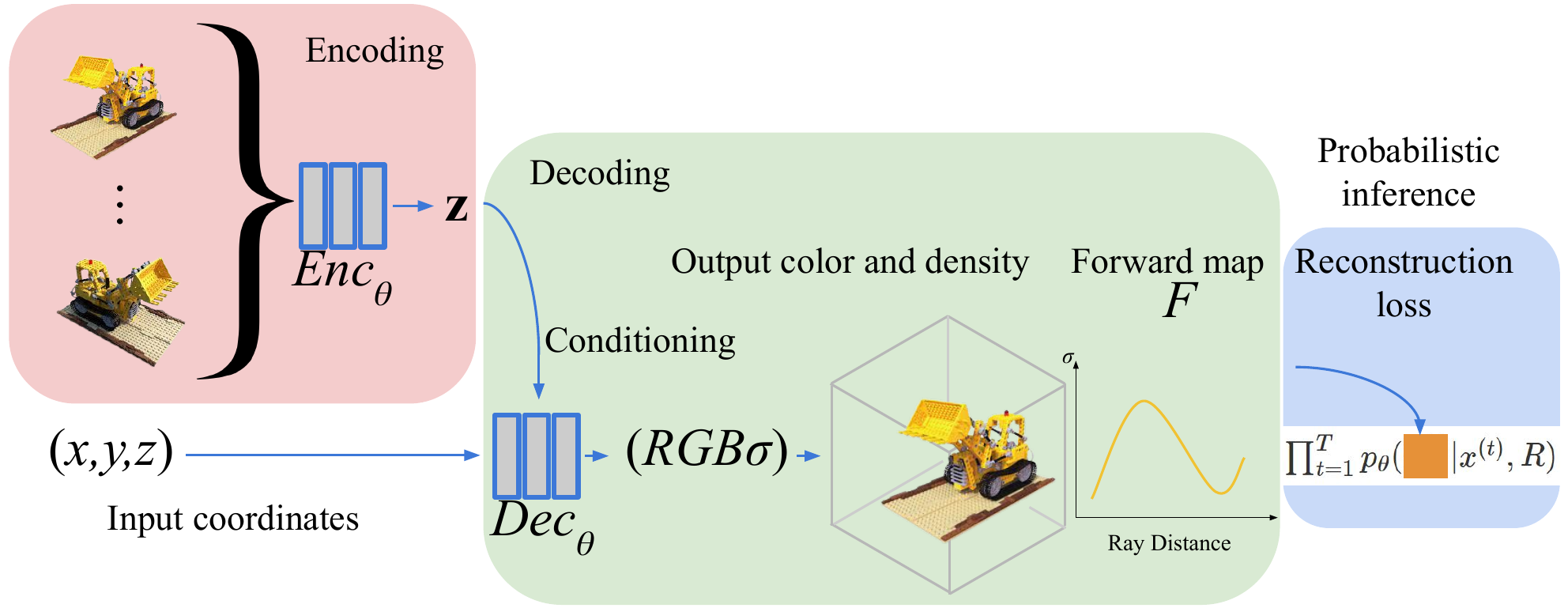}
        \caption{The PONP framework}
        \label{fig:ponp-only-figure}
    \end{subfigure}


\caption{An example of PONP using NeRF \cite{mildenhall2020nerf}. (a) The typical neural field training framework with forward map $F$, which is essentially the bottom branch of (b). (b) Our proposed PONP framwork. First, ground truth partial sensor observations are encoded into a representation $\mathbf{z}$, possibly with sensor parameters (not shown). Second, $\mathbf{z}$ is used to condition a neural field decoder, and we generate a prediction for the field quantities and apply $F$. Third, we supervise the network with probabilistic inference (e.g. maximum likelihood). Unlike (a), the networks are shared across all signals (not shown). }
\label{fig:np-framework}
\end{figure*}

Typically, the field quantities are not themselves directly observable, but partial sensor observations are available. The function relating the field quantities to sensor observations is called the forward map. The most common algorithm for training neural fields is thus follows the following paradigm \cite{xie2022neural}: first, coordinates are sampled and passed through a neural network to predict the corresponding field quantities.  Then the forward map maps the field quantities to the sensor domain, where we can calculate a reconstruction loss between the neural network's predictions and our ground-truth partial observations in the sensor domain. This reconstruction loss is then used to supervise the training of the neural field (see Figure \ref{fig:np-framework}). 

However, one of the major drawbacks of this neural field training paradigm is that it is computationally expensive for a large dataset of signals, since a separate neural network must be trained from scratch for each signal, requiring a large amount of memory and computation \cite{lee2021meta}, which we will henceforth refer to as the neural field generalization problem.  
Previous approaches to this problem include conditioning, hypernetworks \cite{chen2022transformers}, and gradient-based meta-learning \cite{tancik2021learned, lee2021meta}. 
The first two approaches \cite{ galanti2020modularity, sitzmann2020metasdf, sitzmann2020implicit} seek to directly learn some or all of the parameters of a neural field. The third approach views the problem as a meta-learning problem, where each individual signal is considered a different task \cite{sitzmann2020metasdf, tancik2021learned, lee2021meta}, and apply standard gradient-based meta-learning algorithms such as MAML \cite{finn2017model} or Reptile \cite{nichol2018first} to find a parameter initialization that can be quickly fine-tuned for any specific signal. Previous results are mixed: \cite{sitzmann2020metasdf} finds that gradient-based meta-learning can outperform concatenation and MLP hypernetworks, whereas \cite{chen2022transformers} finds that hypernetworks with a vision transformer \cite{dosovitskiy2020image} encoder can outperform gradient-based meta-learning \cite{tancik2021learned} on tasks where the available partial observations are 2D images.

However, gradient based meta-learning suffers from many problems, such as underfitting on large datasets and hyperparameter sensitivity. Recent research \cite{gao2022matters} finds that neural processes, another meta-learning framework, are more flexible and efficient for complex visual regression tasks, which we hypothesize will carry over to the neural field domain. Additionally, one of our key observations is that the encoder-decoder structure of neural processes is an extension of the hypernetwork approach, making a neural process-style framework a natural choice for the neural field generalization problem. However, the usual neural process framework can not be applied directly to most neural field tasks, since in most neural field tasks the training of the neural field is supervised by partial sensor observations, whereas neural processes generally assume that we have direct observations of the field available. Therefore, we propose a simple partially-observed framework to adapt the neural process framework to the neural field domain that incorporates the forward map relating field observations to sensor observations. 


In this work, we propose neural processes as an alternative to the traditional gradient-based meta-learning and hypernetwork approaches for neural field generalization. Our contributions are as follows:
\begin{enumerate}
    \item We propose to use neural processes to tackle the neural field generalization problem. In particular, we adapt the traditional neural process framework to the common neural field setting where only partial observations of the field are available, and our framework is agnostic to the neural process architecture, allowing us to incorporate new advances in neural processes.
    \item We show that neural processes can outperform both state-of-the-art gradient-based meta learning and state-of-the-art hypernetwork approaches on typical neural field generalization benchmarks, including 2D image regression and completion \cite{tancik2021learned, chen2022transformers}, CT reconstruction from sparse views \cite{tancik2021learned}, and recovering 3D shapes from 2D image observations \cite{tancik2021learned, chen2022transformers}. 
\end{enumerate}

\section{Related Works}

\paragraph{Fitting neural fields for multiple signals}
\label{rel-works-gen} 
There are three main approaches to efficiently train neural networks for a large dataset of signals. The first two approaches are latent variable approaches where a latent representation is leveraged to produce the weights of the neural field. The latent code can be learned through an encoder or through auto-decoding \cite{xie2022neural}. The first approach is to concatenate a latent representation of the signal to the input coordinates. This has been shown to be equivalent to defining an affine function $\Psi$ that maps the latent code to the biases of the first layer of the neural representation \cite{dumoulin2018feature, sitzmann2020metasdf, mehta2021modulated, xie2022neural}. Hypernetworks \cite{sitzmann2019scene, sitzmann2020implicit, sitzmann2021light, nirkin2021hyperseg, chiang2022stylizing} generalize the concatenation approach by seeking to predict the complete set of weights of a neural field. The third major approach is to use gradient-based meta-learning in order to learn a prior over the signals, which is specialized to the final neural field for any particular signal by test-time optimization, and previous work has shown this approach to be superior to the hypernetwork approach \cite{sitzmann2020metasdf}. Recent work \cite{chen2022transformers} has returned to the hypernetwork idea by proposing to use vision transformers \cite{dosovitskiy2020image} as the encoder of the hypernetwork, and show that it can outperform gradient-based meta-learning methods. However, this approach is limited to settings where the partial observations or sensor data are 2D images. Neural processes were included as a baseline to the proposed meta-learning method in \cite{sitzmann2020metasdf}, but otherwise has not been explored in the literature. We show that advances in neural process architectures \cite{kim2019attentive, gordon2019convolutional} allow their performance to exceed that of gradient-based meta-learning, and that this persists when moving to more complex forward maps. 

\paragraph{Gradient-based Meta-learning}
\label{maml}
Gradient-based meta-learning algorithms such as MAML \cite{finn2017model} or Reptile \cite{nichol2018first} have been the dominant approach \cite{sitzmann2020metasdf, tancik2021learned, lee2021meta} used to efficiently train neural fields over many signals. This approach takes the view that for a dataset of neural fields, each neural field is a specialization of a meta-network \cite{xie2022neural}. \cite{sitzmann2020metasdf} finds that using the gradient-based meta-learning of neural fields for signed-distance functions outperform concatenation, hypernetworks, and conditional neural processes. 
\cite{tancik2021learned} applies standard gradient-based meta-learning algorithms (MAML and Reptile) to image regression, CT reconstruction, and view synthesis tasks. Our work instead proposes to use a new partially-observed NP framework to perform these tasks. \cite{lee2021meta} proposes using MAML in conjunction with pruning techniques to learn sparse neural representations. This method tackles a problem that is orthogonal to the generalization problem and could possibly be adapted to work in conjunction with our method.

\paragraph{Neural Processes}
\label{rel-works-nps}

Neural processes is a meta-learning framework that aims to learn a distribution over models by modelling a stochastic process. Architecturally, Neural processes are encoder-decoder models that are trained with probabilistic inference. There are two major classes of neural processes: conditional neural processes \cite{garnelo2018conditional, dubois2020npf, gordon2019convolutional} and latent neural processes \cite{garnelo2018neural, kim2019attentive, foong2020meta}, which are latent variable models. Latent variable neural processes can offer better modelling at the cost of having intractable training objectives \cite{dubois2020npf}. There are three major architectures for neural processes: MLP \cite{garnelo2018conditional, garnelo2018neural}, where both the encoder and decoder are MLPs and the an aggregate representation is produced via taking the mean, attention-based \cite{kim2019attentive}, where the aggregator is an attention module, and convolutional \cite{gordon2019convolutional, foong2020meta}, which uses a discrete version of the convolution operator in order to give the network translation equivariance. MLP-based models tend to underfit compared to the other two classes of models, and convolutional models tend to create smoother samples compared to attentive models \cite{kim2019attentive, dubois2020npf}. Neural processes have been used for low-dimensional regression \cite{garnelo2018neural, garnelo2018conditional, kim2019attentive, gordon2019convolutional, foong2020meta}, image completion \cite{garnelo2018neural, garnelo2018conditional, kim2019attentive, gordon2019convolutional, foong2020meta}, Bayesian optimization \cite{garnelo2018neural, foong2020meta}, and visual regression tasks \cite{gao2022matters}. There is concurrent work \cite{guo2023versatile} that proposes a Versatile Neural Process (VNP) architecture, based on attentive neural processes \cite{kim2019attentive}, for neural field generalization. In contrast, our work proposes a neural process framework that is both encoder and neural-field agnostic. 

\section{Method}

In this section, we will first introduce necessary notation and give an overview of neural processes (Section \ref{neural-processes-intro}). We then relate neural processes to neural fields and present our proposed neural process-based framework for neural field generalization (Section \ref{np-framework}). Finally, we will describe the training of our neural process framework in Section \ref{prob-inference}.


\paragraph{Notation}
In the meta-learning setting, we have a set of tasks $\mathcal{M} = \{\mathcal{D}_i\}_{i = 1}^{N_\text{tasks}}$. Each task $\mathcal{D}_i$ is a dataset with a training set, which we will call the \textbf{context set} and denote by $\mathcal{C}$, and a test set, which consists of \textbf{target inputs} (denoted $\mathbf{x}_\mathcal{T}$) and \textbf{target outputs} (denoted $\mathbf{y}_\mathcal{T})$. Our goal is to train a function $f(\cdot; \mathcal{C})$ such that given the context $\mathcal{C}$ of some new task, we get a predictor $f_\mathcal{C}(\cdot)$ that maps the target inputs to the target outputs with low error. 

When discussing neural fields, we follow the terminology of \cite{xie2022neural}: data points or signals are viewed as a \textbf{field} $\Phi: \mathcal{X} \to \mathcal{Y}$. that assigns to all coordinates $\mathbf{x} \in \mathcal{X}$ some quantity $\mathbf{y} \in \mathcal{Y}$. A \textbf{neural field} parameterizes $\Phi$ with a neural network with parameters $\Theta$, which we will denote $\Phi_\Theta$. The goal is to find $\Theta$ such that $\Phi_\Theta(\mathbf{x}) \approx \Phi(\mathbf{x)}, \forall \mathbf{x} \in \mathcal{X}$.” As discussed above, typically we only have partial observations from a sensor domain. The sensor data is related to the field quantities via a \textbf{forward map}, which we formally view as an operator $F(\Phi; \theta)$ that takes in a neural field $\Phi$ and sensor parameters $\theta$ and produces an output in the sensor domain, which is typically much lower-dimensional. Examples of sensor parameters $\theta$ are the projections angles in CT reconstruction or the camera parameters for NeRFs. We will assume that $F$ is differentiable, allowing the use of gradient descent optimization.  

\subsection{Neural Processes} 
\label{neural-processes-intro}
Neural processes \cite{garnelo2018neural} are a class of latent-variable neural network models that aim to learn distributions over functions. This makes them a natural meta-learning algorithm, as they can be viewed as a probabilistic model over task predictors. Neural processes seeks to parameterize the distribution $p(\mathbf{y}_\mathcal{T} | \mathbf{x}_\mathcal{T}, \mathcal{C})$ with a neural network. The basic structure of a neural process consists of three main components \cite{garnelo2018neural}: 1.) a permutation-invariant encoder that encodes the context set $\mathcal{C}$, 2.) an aggregator that produces a representation $\mathbf{z}$ of $\mathcal{C}$, and 3.) a decoder that takes as input both the representation $\mathbf{z}$ as well as the target inputs $\mathbf{x}_i$ and outputs $p(\mathbf{y}_\mathcal{T} | \mathbf{x}_\mathcal{T}, \mathcal{C})$. Training a NP proceeds as follows: first a dataset $\mathcal{D}_i \in \mathcal{M}$ is chosen, and then a context (training) set $\mathcal{C}$ is chosen from $\mathcal{D}_i$. $\mathcal{C}$ is then encoded into a global representation $\mathbf{z} = A(E(\mathbf{x}_\mathcal{C}, \mathbf{y}_\mathcal{C}))$. The decoder takes as input the test inputs $\mathbf{x}_\mathcal{T}$, is conditioned with $\mathbf{z}$, and outputs $p(\mathbf{y}_\mathcal{T}|\mathbf{x}_\mathcal{T}, \mathcal{C}) = D(\mathbf{x}_\mathcal{T}; \mathbf{z})$. The true log likelihood $\mathcal{L} = \log p_\theta(\mathbf{y}_\mathcal{T}|\mathbf{x}_\mathcal{T}; \mathcal{C})$ is then computed analytically or estimated. The gradient $\nabla \mathcal{L}$ is used to optimize the parameters of the NP via stochastic gradient optimization.  

\paragraph{Conditional NPs vs Latent NPs}
There are two main ways to model the distribution $p(\mathbf{y}_\mathcal{T} | \mathbf{x}_\mathcal{T}, \mathcal{C})$: the first assumes conditional independence with respect to $\mathcal{C}$ and are known as \textbf{conditional neural processes (CNPs)}, and the second, which models the distribution with a probabilitic latent vector $\mathbf{z}$ in the same manner as a variational autoencoder (VAE), known as \textbf{latent neural processes (LNPs)}. Implementation-wise, CNPs resemble autoencoder whereas LNPs resemble VAEs. CNPs can be trained with maximum likelihood estimation, but for LNPs, as with VAEs, the true likelihood is intractable. To get around this, the true likelihood can be estimated either using Monte-Carlo integration, called NPML \cite{foong2020meta}, or with variational inference (NPVI) \cite{garnelo2018neural}. LNPs trained with NPML often outperform those trained with NPVI but require more computation in the form of additional Monte-Carlo samples \cite{foong2020meta}.


\subsection{Our Partially-Observed Neural Process Framework}
\label{np-framework}

\paragraph{Connection between neural processes and neural fields} 
To illustrate the connection between neural processes and neural fields, we examine one of the tasks frequently benchmarked in the neural process literature: image completion \citep{garnelo2018neural, garnelo2018conditional, kim2019attentive, gordon2019convolutional, foong2020meta}, where a subset of the pixels of an image and their color values are provided as the context inputs and outputs, and the goal is to predict the color values of all the pixels of the image. The target inputs are the coordinates of pixels, and the desired target outputs are the color values of those pixels. Our key observation is that this is a neural field task, with the forward map being the projection map onto the seen context pixels. Recall that in the neural process framework (Section \ref{neural-processes-intro}), the decoder of the neural process takes as input coordinates as well as a representations of the context set, and returns their color values, so the neural process decoder is equivalent to a conditional neural field (Figure \ref{fig:np-framework}).  In this example, the neural process encoder also encodes the partial sensor observations (output of the forward map). 

\paragraph{Our neural process framework}
Based on the observations of the previous section, we propose our partially-observed neural process framework (PONP) for training neural fields (see Figure \ref{fig:np-framework}): 1.) a task-specific encoder, which takes as input partial sensor observations and aggregates them into a representation $\mathbf{z}$, (red box) 2.) a decoder, which consists of a conditional neural field decoder, which is conditioned on $\mathbf{z}$ and designed to take as input field coordinates, and whose output is fed through the forward map $F$, producing a distribution (green box), and 3.) is trained with probabilistic inference (blue box). Note that with our framework we are able to leverage both existing neural process architectures (MLP, attention-based, and convolutional) and latent variables approaches (i.e., LNPs). Details on how we define the distribution are deferred to Sec. \ref{prob-inference}.

\paragraph{Conditioning the neural field}
In order to produce different neural fields for different input signals, we need to condition our neural field decoder using the learned representation extracted from our context set. The simplest approach for doing this is to concatenate the representation with the input to the neural field \cite{sitzmann2020metasdf}. However, since concatenation is equivalent to predicting the biases of the first layer of a neural network, concatenation is only suitable for datasets with very similar signals, such as the 2D CT reconstruction dataset \cite{tancik2021learned}. To learn more diverse signals, more powerful approaches are required. Hypernetworks \cite{ha2016hypernetworks} generalize concatenation by having the encoder learn to produce the complete set of  weights for the neural field, at the cost more parameters. This approach has been used by previous methods \cite{sitzmann2020metasdf, chen2022transformers, sitzmann2021light} as a way to learn more complex structures. Our choice of conditioning varies based on the task structure.  

\subsection{Probabilistic inference} 
\label{prob-inference}
Recall that neural processes are trained with a maximum likelihood objective. For our neural process-inspired framework, there are two choices of how to define the distribution: either over the field space or over the sensor space. As discussed previously, these spaces are typically not the same, as the sensor space is generally lower dimensional. Since we only have groundtruth observations in the sensor domain, defining our distribution over the field space has the disadvantage that we need to use the change-of-variables formula for probability distributions on the forward map $F$ to compute the likelihood of groundtruth observations as required by maximum likelihood training. Since the forward map $F$ may be very complex and thus not bijective or even injective, we cannot use the typical (bijective) change-of-variables formula
but instead need to use a more general change-of-variables formula, which are much harder to compute or estimate. Defining the distribution on the sensor space avoids these problems.

\paragraph{Our framework}
In our framework, we define our distribution as an independent Gaussian distribution at each coordinate in the sensor domain. Practically, this is implemented as follows: given a conditional neural field as the first stage of the decoder, we modify the last layer to have two heads, producing outputs $M, \Sigma$. The output of both heads are then fed separately through the forward map $F$ to produce $\mu, \sigma$, which become the mean and standard deviation of our distribution, respectively. 


\section{Experiments}

\subsection{Experimental Setup}

We adapt the following experimental setup, which consists of two steps: 1.) meta-learning and 2.) test-time optimization, if necessary. Meta-learning consists of training a model on all signals in order to find a good initialization for the neural fields, and test-time optimization specializes the weights of the neural field to a particular signal. In order to make a fair comparison, we allow the same number of test-time optimization steps, if necessary, for all methods, and use comparable neural field architectures. For all tasks, reconstruction quality is measured by peak signal-to-noise ratio (PSNR). 

\subsection{Baselines}

We use the following baselines: standard neural field initialization, with test-time optimization, the method of \cite{tancik2021learned}, which applies a standard gradient-based meta-learning algorithm (either MAML \cite{finn2017model} or Reptile \cite{nichol2018first}) and then fine-tunes with test-time optimization, and where applicable the Transformer INR method \cite{chen2022transformers}, which uses a vision transformer-based hypernetwork approach that uses a vision transformer to learn the weights of a neural field from 2D image sensor data. The use of a vision transformer encoder means that the Transformer INR method is not suitable for all tasks. Unlike the previous two baselines, this method does not require test-time optimization. 


\subsection{Tasks}
In order to directly compare our neural process approach against previous methods, we test our method using four benchmarks, three of which were first proposed by \cite{tancik2021learned}. More details about the tasks can be found in Appendix \ref{experiment-deets}.

\begin{table}[h]
\caption{Forward maps $F$ for our three tasks.}
\vspace{-1em}
\resizebox{\columnwidth}{!}{
\begin{tabular}{lll}
    \toprule
        Task & $F$ & Input \& Output \\
    \midrule
    Image & Masking & Pixel Location $\rightarrow$ RGB value \\
    CT & Integral projection & 2D image $\rightarrow$ 1D sinogram \\
    NVS & Volume rendering & Radiance field $\rightarrow$ 2D image \\
    \bottomrule
\end{tabular}
}
\vspace{-1em}
\label{table:forward-maps}
\end{table}

\paragraph{2D image regression and completion}
In the image regression task, a neural field is trained to represent a 2D image by taking as input 2D pixel coordinates and outputting the corresponding RGB color values \cite{tancik2021learned}. We use the CelebA \cite{liu2015deep} dataset, which consists of over 200K images of human faces. We downsample each image to $32 \times 32$. For the 2D image completion task, the groundtruth RGB values are known at some percentage of the pixels and the rest of the pixels are masked. As discussed earlier, this is a neural field task where the forward map is the indicator function on the pixels for which we have groundtruth (see Table \ref{table:forward-maps}).

\begin{table}[ht!]
\caption{Comparison of neural field generalization methods on the 2D image regression task. The best PSNRs in the whole table are shown in bold and the second-best PSNRs are underlined. Parameters refers to the total number of parameters for the model, including the neural field itself. The TTO column is checked if test-time optimization was used.}
\vspace{-0.4cm}
\begin{center}
\begin{small}
\begin{sc}
\begin{tabular}{lccc}
    \toprule
    Model & TTO & PSNR & Params \\
    \midrule
    Random Init. & \: \cmark & 10.87 & 50K\\
    Reptile \cite{tancik2021learned} & \: \cmark & 25.27 & 50K\\
    MAML \cite{tancik2021learned} & \: \cmark & 32.36 & 50K\\
    Transformer INR \cite{chen2022transformers} & \: \xmark & \underline{48.36} & 44.3M\\
    \midrule
    \textbf{PONP (Ours)} & \: \xmark & \textbf{78.87} & 340K \\
    \bottomrule
\end{tabular}
\end{sc}
\end{small}
\end{center}
\vskip -0.1in
\label{table:img-reg}
\vspace{-0.4cm}
\end{table}

\begin{table}[htb]
\caption{Comparison of neural field generalization methods on the 2D image completion task. The best PSNRs in the whole table are shown in bold and the second-best PSNRs are underlined. Parameters refers to the total number of parameters for the model, including the neural field itself. The TTO column is checked if test-time optimization was used.}
\begin{center}
\begin{small}
\begin{sc}
\vspace{-0.4cm}
\begin{tabular}{lccc}
    \toprule
    Model & TTO & PSNR & Params \\
    \midrule
    Random Init. & \: \cmark & 10.23 & 50K\\
    Reptile \cite{tancik2021learned} & \: \cmark & 14.65 & 50K\\
    MAML \cite{tancik2021learned} & \: \cmark & 13.35 & 50K\\
    Transformer INR \cite{chen2022transformers} & \: \xmark & \underline{18.09} & 44.3M \\
    \midrule
    \textbf{PONP (Ours)} & \: \xmark & \textbf{23.24} & 340K \\
    \bottomrule
\end{tabular}
\end{sc}
\end{small}
\end{center}
\vskip -0.1in
\label{table:img-comp}
\vspace{-0.4cm}
\end{table}

\begin{figure*}[ht!]
\begin{center}
    \begin{tabular}{ccccc|c}
    \rotatebox{90}{Reptile} & \includegraphics[width=0.12\linewidth]{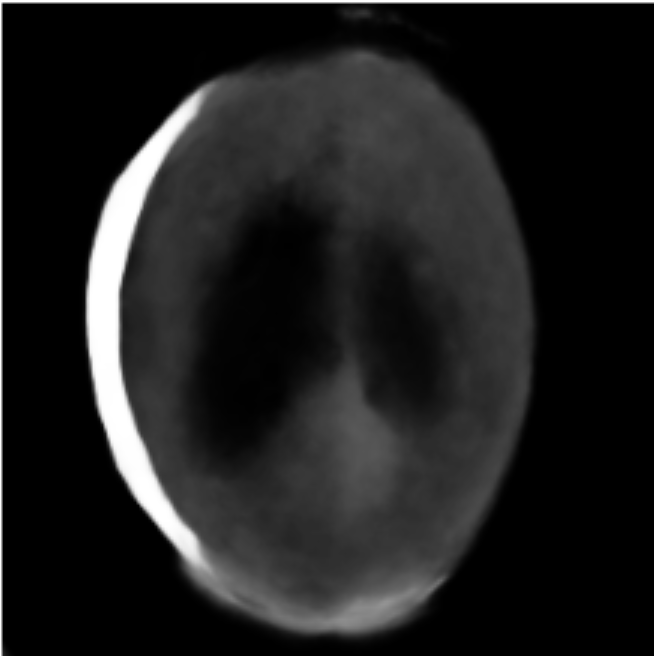} & \includegraphics[width=0.12\linewidth]{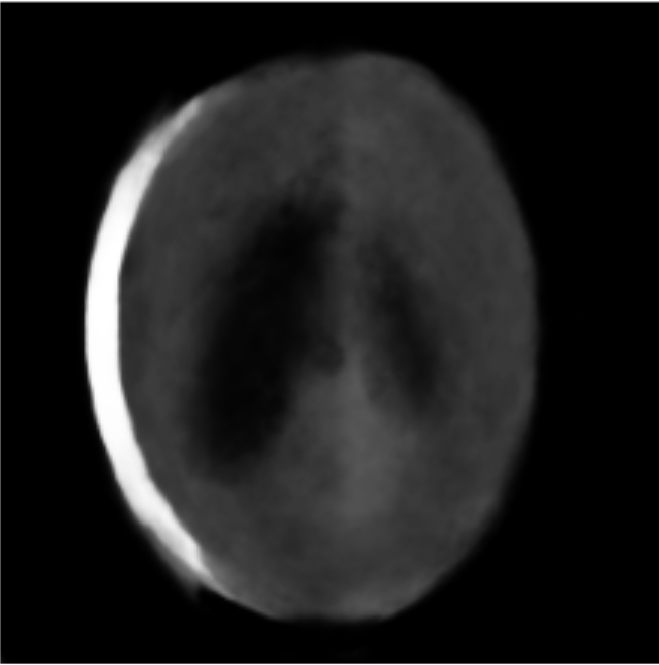} & \includegraphics[width=0.12\linewidth]{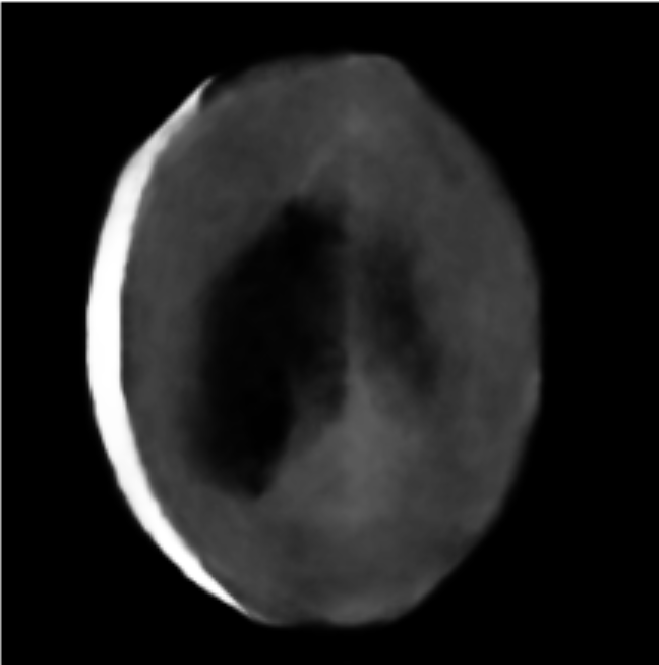} & \includegraphics[width=0.12\linewidth]{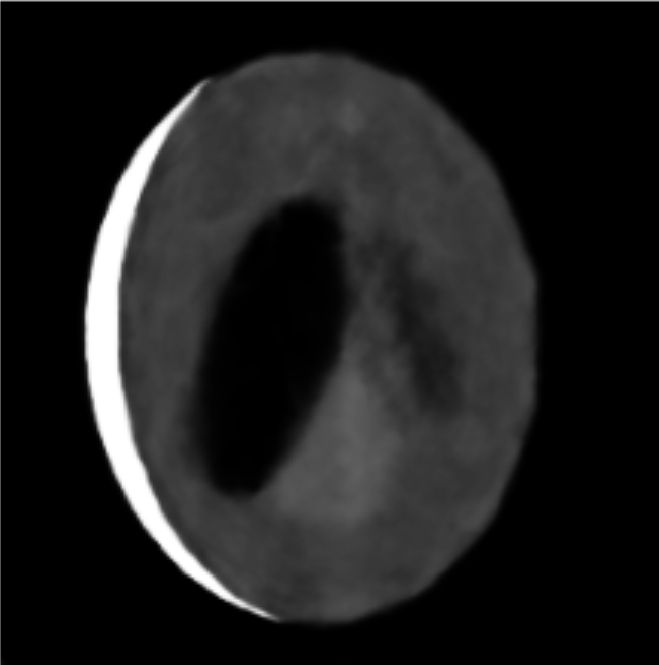} &  \includegraphics[width=0.12\linewidth]{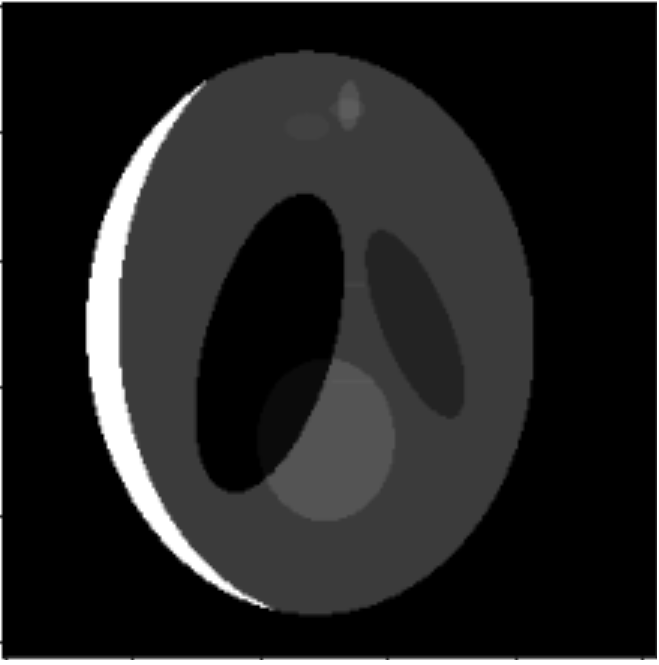} \\
    \rotatebox{90}{NP} & \includegraphics[width=0.12\linewidth]{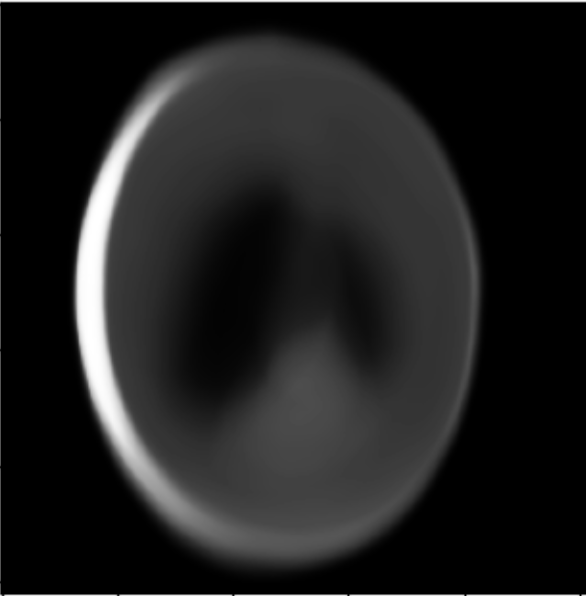} & \includegraphics[width=0.12\linewidth]{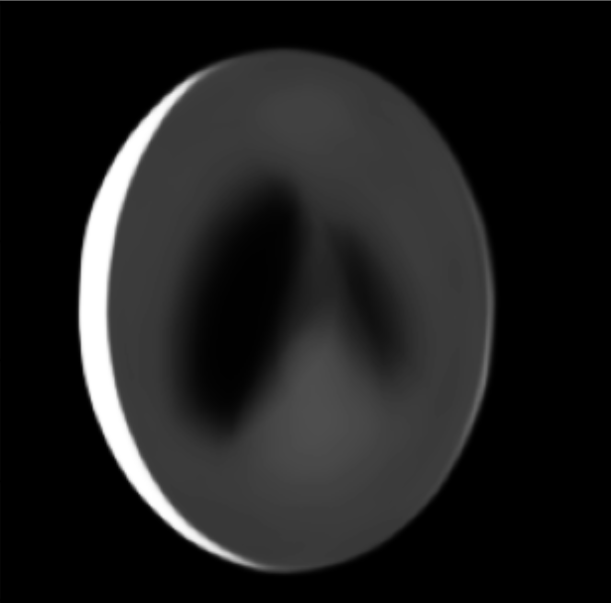} & \includegraphics[width=0.12\linewidth]{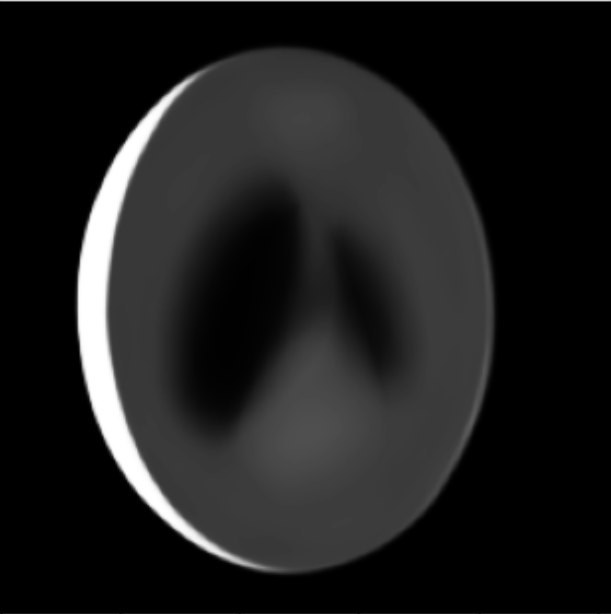} & \includegraphics[width=0.12\linewidth]{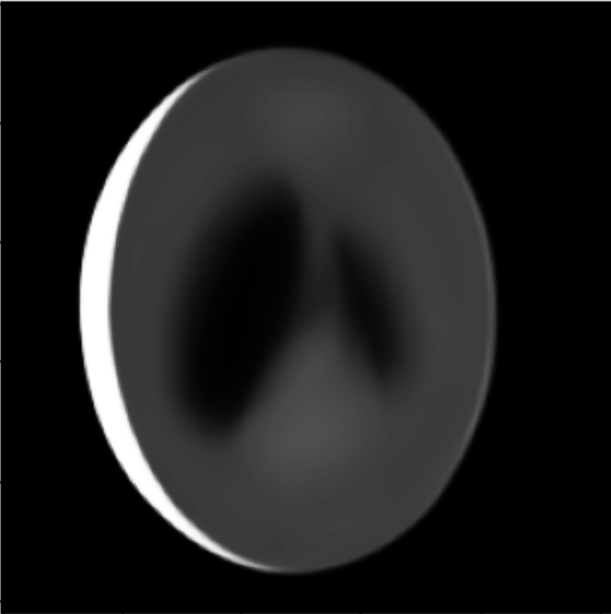} & \includegraphics[width=0.12\linewidth]{images/ct_gt_new.png} \\
    \rotatebox{90}{NP (TTO)} & \includegraphics[width=0.12\linewidth]{images/np_view_1.png} & \includegraphics[width=0.12\linewidth]{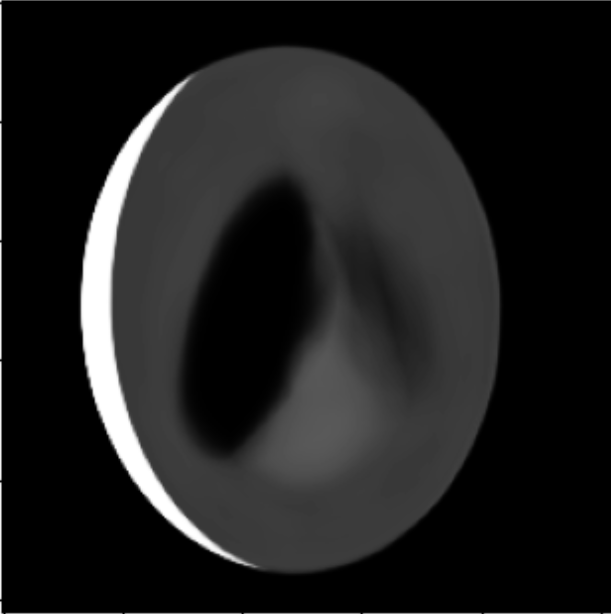} & \includegraphics[width=0.12\linewidth]{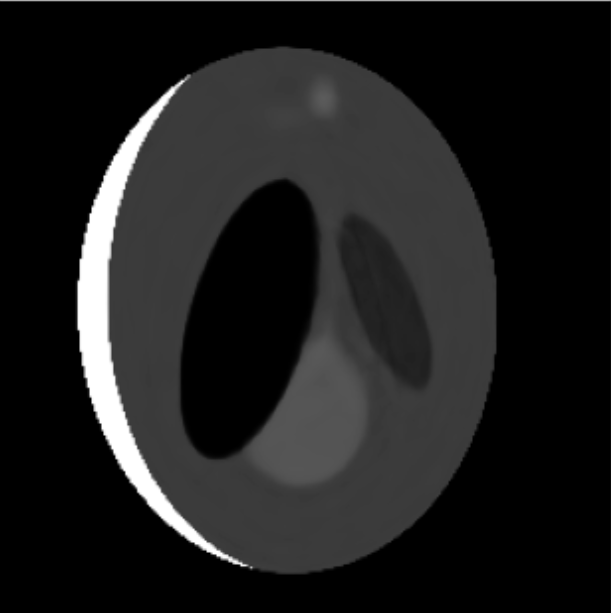} & \includegraphics[width=0.12\linewidth]{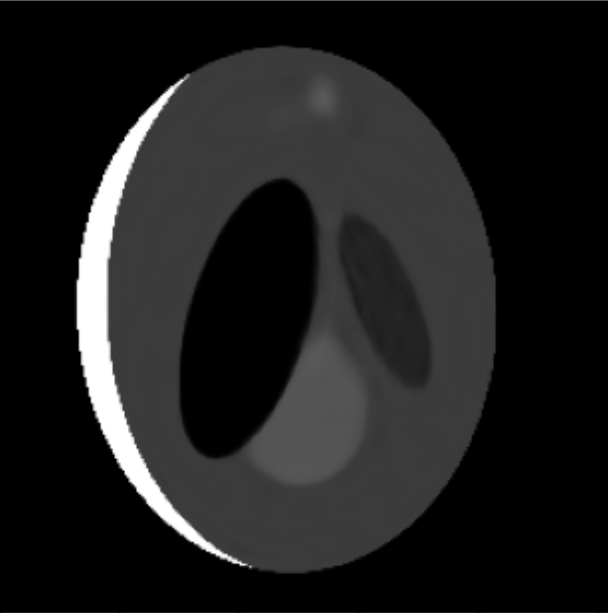} & \includegraphics[width=0.12\linewidth]{images/ct_gt_new.png} \\
     & 1 View & 2 Views & 4 Views & 8 Views & Target \\
\end{tabular}
\end{center}
\caption{Examples of CT scan reconstruction from a sparse set of projections. The reconstructed scans learned by the neural process models are much sharper than their counterparts generated by gradient-based meta-learning.}
\label{fig:ct}
\end{figure*}

\begin{table*}[htb]
\caption{Comparison of initialization methods on the CT Reconstruction task. The best PSNRs in the whole table are shown in bold and the second-best PSNRs are underlined. The TTO column is checked if test-time optimization was used. }
\vskip 0.15in
\begin{center}
\begin{small}
\begin{sc}
\begin{tabular}{lccccc}
    \toprule
    Model & TTO & 1 Views & 2 Views & 4 Views & 8 Views \\
    \midrule
    Random Init. \cite{tancik2021learned} & \: \cmark & 13.63 & 14.15 & 16.31 & 21.49 \\
    Reptile \cite{tancik2021learned} & \: \cmark & 15.09 & 18.70 & 22.00 & \underline{27.34} \\
    \textbf{PONP (Ours)} & \: \xmark & \underline{16.39} & \underline{20.39} & \underline{22.13} & 22.60 \\
    \textbf{PONP (Ours)} & \: \cmark & $\mathbf{24.67}$ & $\mathbf{29.11}$ & $\mathbf{37.54}$ & $\mathbf{37.66}$ \\ 
    \bottomrule
\end{tabular}
\end{sc}
\end{small}
\end{center}
\vskip -0.1in
\label{table:ct-comparison}
\vspace{-0.4cm}
\end{table*}

\paragraph{2D CT reconstruction} The CT reconstruction task is to reconstruct a 2D CT scan given a sparse set of 1D projections of the CT scan. The CT reconstruction dataset consists of 2048 randomly generated $256 \times 256$ Shepp-Logan phantoms \cite{shepp1974fourier}. The projections are generated from the CT scans by the forward map $F$ given by 2D integral projections of a bundle of 256 parallel rays from a given angle (see Table \ref{table:forward-maps}). At test time, we are given 1, 2, 4, and 8 projections (views) with which to reconstruct the phantom. We are allowed 50, 100, 1000, and 1000 test-time optimization steps, respectfully, for each of the four settings.

\paragraph{View synthesis for ShapeNet objects} View synthesis is the task of generating a novel view of a 3D object or scene given some number of input views, and is typically tackled with neural radiance fields \cite{mildenhall2020nerf, chen2022transformers}. Neural radiance fields learn the radiance of a 3D scene with a neural field and use the volume rendering forward map to generate 2D views from a given viewing angle (see Table \ref{table:forward-maps}). For the ShapeNet \cite{chang2015shapenet} view synthesis task, the data is created from objects taken from one of three ShapeNet classes: \textit{chairs}, \textit{cars}, and \textit{lamps}. For each 3D object, 25 $128 \times 128$ views are generated from viewpoints chosen randomly on a sphere. The goal is to generate given unseen views. The two different settings proposed in \cite{tancik2021learned} are the multi-view setting, where we are allowed to train with the full 25 training views of each object, and the single-view setting, where we are only allowed to train with a single view per object. Since the Transformer INR baseline only reports results on the single-view and 2-shot (2 view) versions of the task, we only consider these tasks as well. We are allowed 1000-2000 test-time optimization steps, depending on the class of the object and the setting (multi-view or single-view). 


\subsection{Results}

\paragraph{2D image regression and completion}
For both the 2D image regression and 2D image completion tasks, the field space and sensor space are the same, so our PONP framework coincides with the normal neural process framework. We use a convolutional CNP architecture \cite{dubois2020npf}, as the convolutional encoder is well-suited to handling 2D image data. For both the image regression and image completion tasks, we set the context and target inputs to be the 2D pixel coordinates, and the context and target outputs as RGB color values. For the image completion task, we randomly mask between 10-30\% of the pixels. For the Transformer INR baseline, which requires 2D images as input, the masked pixels are assigned a value of $\mathbf{0}$. 

Quantitative results for the image regression and completion tasks can be found in Tables \ref{table:img-reg} and \ref{table:img-comp}, respectively. We find that even without the benefit of test-time optimization, our PONP framework greatly outperform gradient-based meta-learning methods and hypernetwork methods. In particular, PONP greatly outperforms the state-of-the-art Transformer INR hypernetwork method while using just a fraction of the parameters. We also observe that, as in \cite{chen2022transformers}, the Transformer INR method outperforms gradient-based meta-learning methods on both tasks. We also find that within gradient-based optimization methods, Reptile outperforms MAML in partially-observed settings, whereas the reverse is true in the fully-observed setting, as in \cite{tancik2021learned}. Qualitative results can be found in the Appendix (Figure \ref{fig:2D-img}).

\begin{figure*}[t]
\begin{center}
    \begin{tabular}{c | cc c| cc cc | cc}
        Input & GT & Pred. & Input & GT & Pred. & \multicolumn{2}{c}{Input} & GT & Pred. \\
        \includegraphics[width=0.08\textwidth]{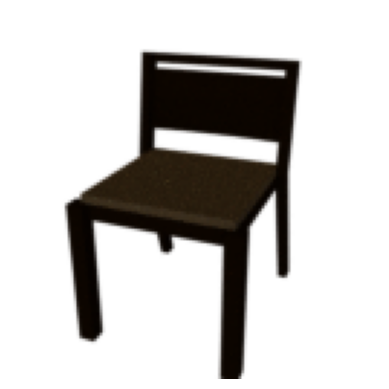} & \includegraphics[width=0.08\textwidth]{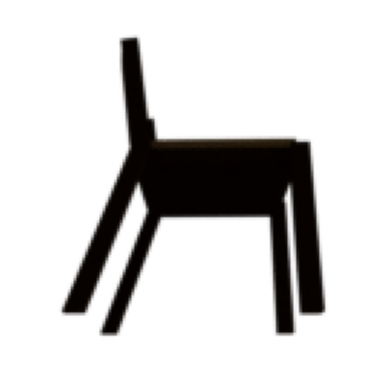} & \includegraphics[width=0.08\textwidth]{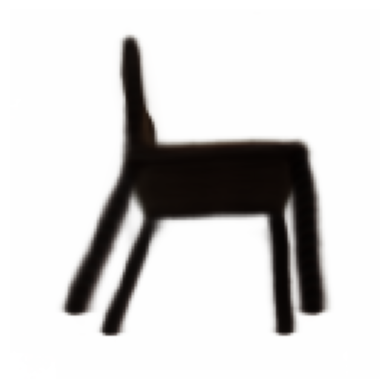} & \includegraphics[width=0.08\textwidth]{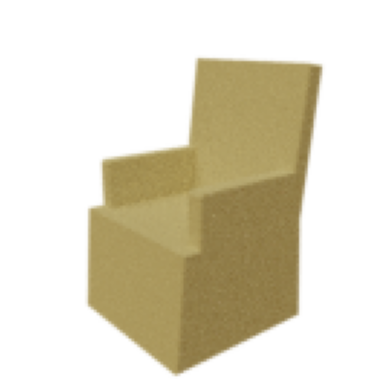} & \includegraphics[width=0.08\textwidth]{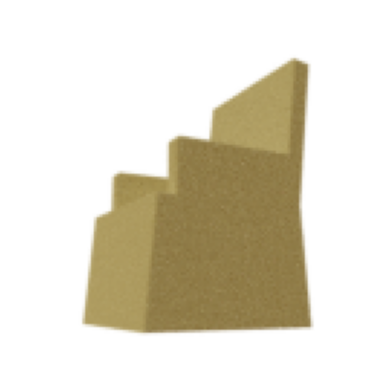} & \includegraphics[width=0.08\textwidth]{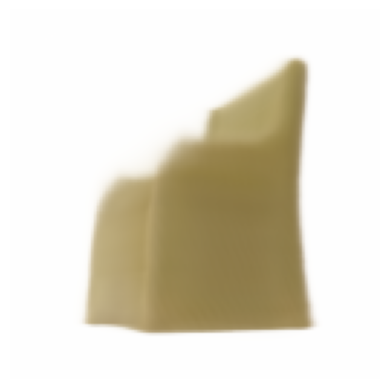} & \includegraphics[width=0.08\textwidth]{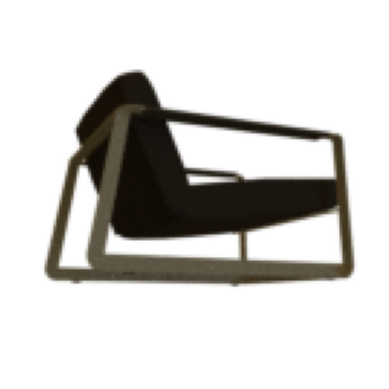} & \includegraphics[width=0.08\textwidth]{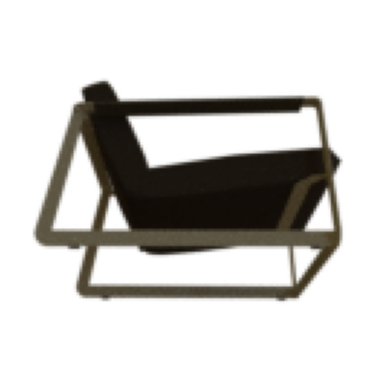} & \includegraphics[width=0.08\textwidth]{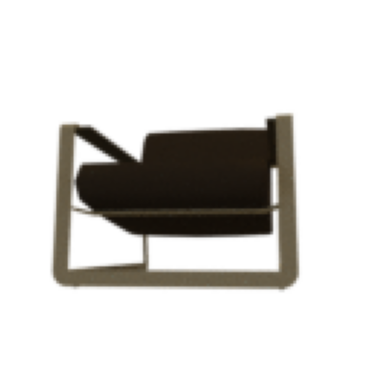} & \includegraphics[width=0.08\textwidth]{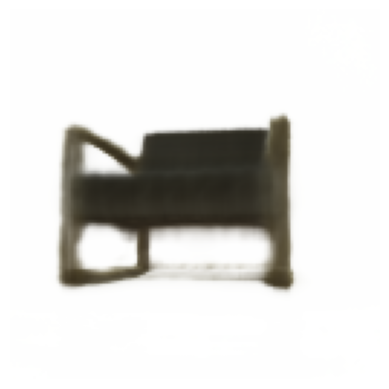} \\
        \includegraphics[width=0.08\textwidth]{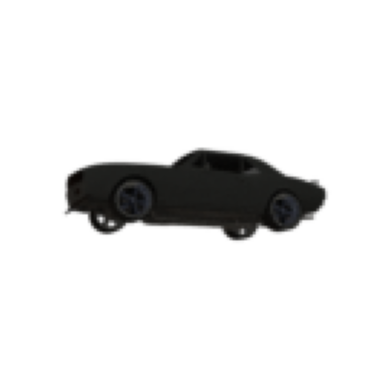} & \includegraphics[width=0.08\textwidth]{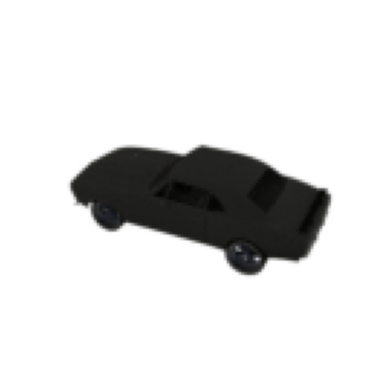} & \includegraphics[width=0.08\textwidth]{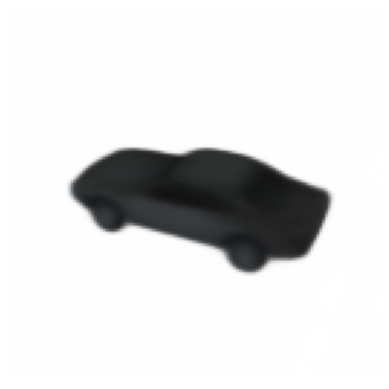} & \includegraphics[width=0.08\textwidth]{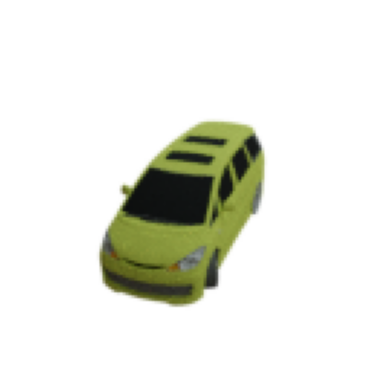} & \includegraphics[width=0.08\textwidth]{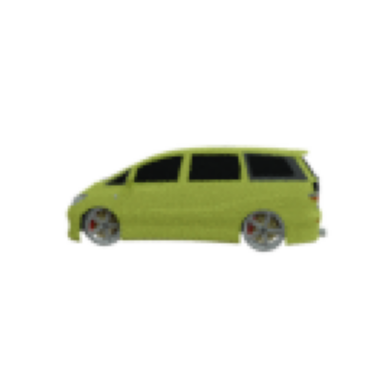} & \includegraphics[width=0.08\textwidth]{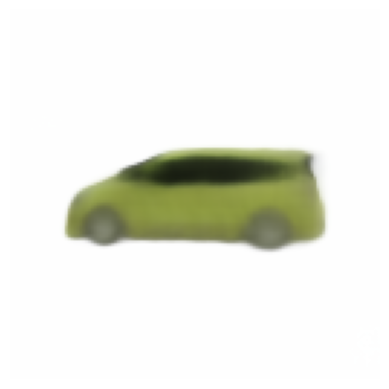} & \includegraphics[width=0.08\textwidth]{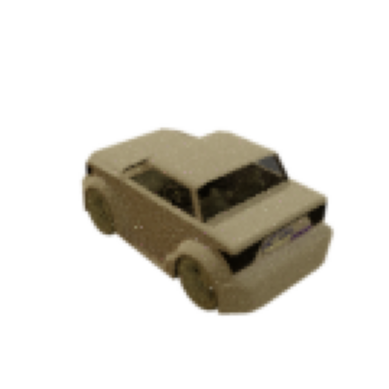} & \includegraphics[width=0.08\textwidth]{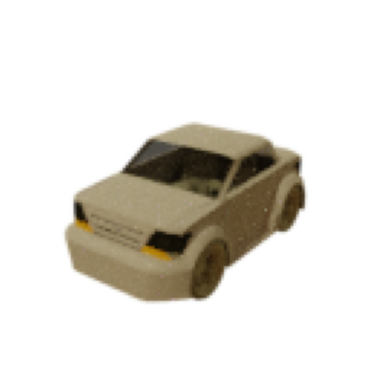} & \includegraphics[width=0.08\textwidth]{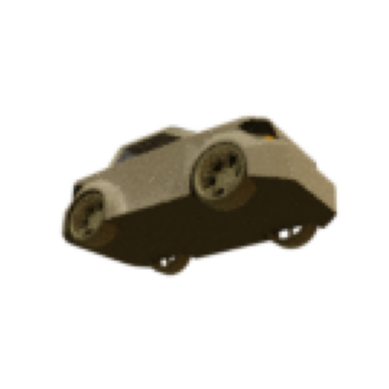} & \includegraphics[width=0.08\textwidth]{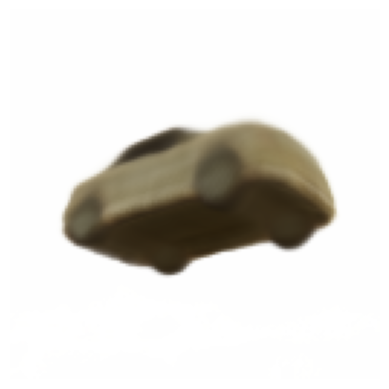} \\
        \includegraphics[width=0.08\textwidth]{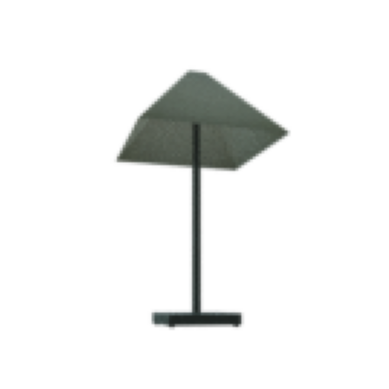} & \includegraphics[width=0.08\textwidth]{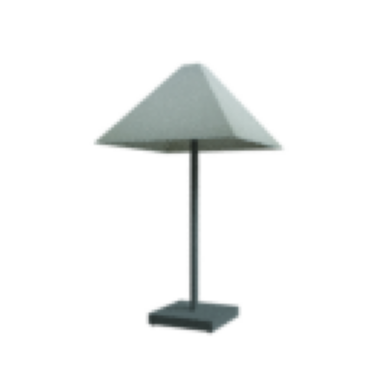} & \includegraphics[width=0.08\textwidth]{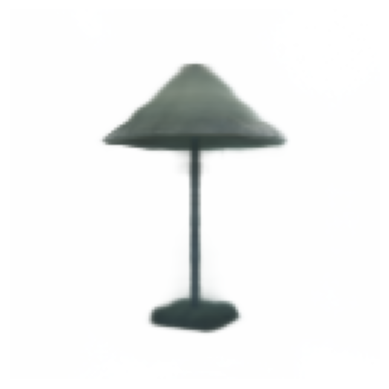} & \includegraphics[width=0.08\textwidth]{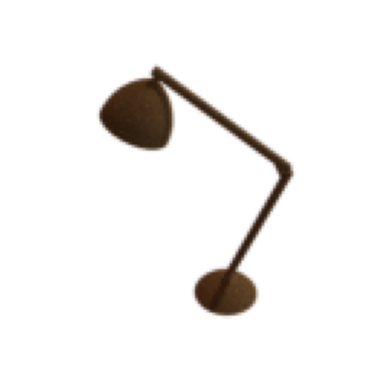} & \includegraphics[width=0.08\textwidth]{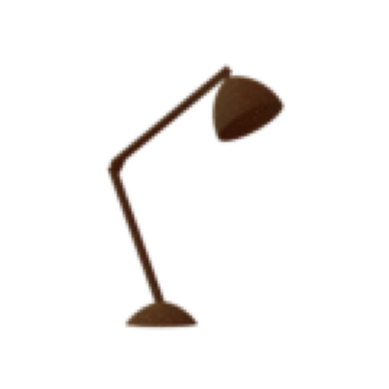} & \includegraphics[width=0.08\textwidth]{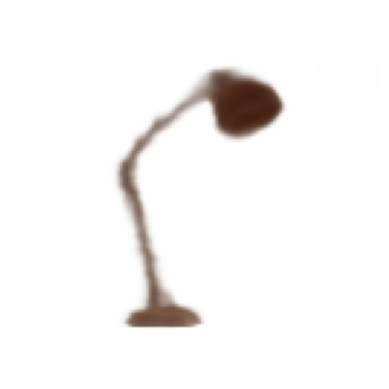} &  \includegraphics[width=0.08\textwidth]{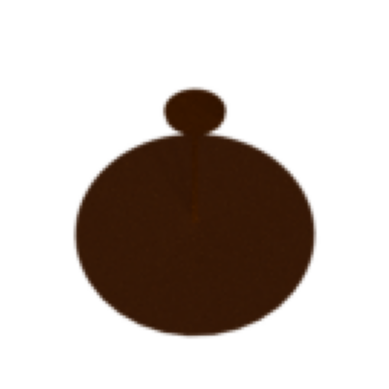} & \includegraphics[width=0.08\textwidth]{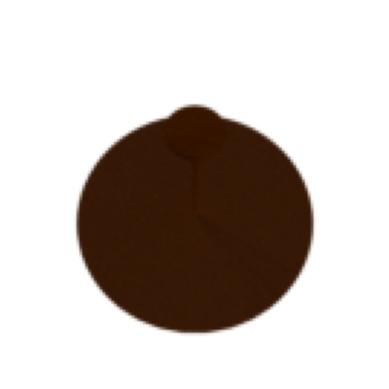} & \includegraphics[width=0.08\textwidth]{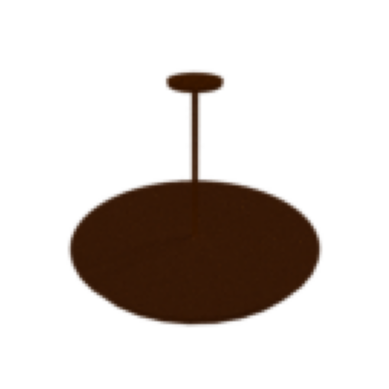} & \includegraphics[width=0.08\textwidth]{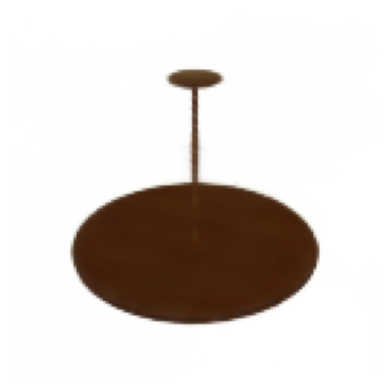} \\
    \end{tabular}
\end{center}
\caption{Qualitative examples of our PONP framework for the ShapeNet view synthesis experiment.}
\label{fig:shapenet}
\end{figure*}

\begin{table*}[t!]
\caption{Comparison of neural field generalization and initialization methods on the ShapeNet view synthesis task. The best PSNRs without test-time optimization and with test-time optimization are bolded. 
}
\vskip 0.15in
\begin{center}
\begin{small}
\begin{sc}
\begin{tabular}{lcccccc}
    \toprule
    Model & Views & TTO & Chairs & Cars & Lamps \\
    \midrule
    Random Init. \cite{tancik2021learned} & 0 & \cmark & 12.49 & 11.45 & 15.47 \\
    Matched \cite{tancik2021learned} & 25 & \cmark & 16.40 & 22.39 & 20.79 \\
    Shuffled \cite{tancik2021learned} & 25 & \cmark & 10.76 & 11.30 & 13.88 \\
    Reptile \cite{tancik2021learned} & 1 & \cmark & 16.54 & 22.10 & 20.95 \\
    Reptile \cite{tancik2021learned} & 25 & \cmark & 18.85 & 22.80 & 22.35 \\
    \midrule
    Transformer INR 
    \cite{chen2022transformers} & 1 & \xmark & \textbf{19.66} & 23.78 & 22.76 \\
    \textbf{PONP (Ours)} & 1 & \xmark & 19.48 & \textbf{24.17} & \textbf{22.78} \\
    \midrule
    Transformer INR 
    \cite{chen2022transformers} & 2 & \xmark & 21.10 & 25.45 & 23.11 \\
    \textbf{PONP (Ours)} & 2 & \xmark & \textbf{21.13} & \textbf{25.98} & \textbf{23.28} \\
    \midrule
    Transformer INR \cite{chen2022transformers} & 1 & \cmark & \textbf{20.56} & 24.73 & 24.71 \\
    \textbf{PONP (Ours)} & 1 & \cmark & 20.55 & \textbf{24.99} & \textbf{24.86}  \\
    \midrule
    Transformer INR 
    \cite{chen2022transformers} & 2 & \cmark & 23.59 & 27.13 & 27.01 \\
    \textbf{PONP (Ours)} & 2 & \cmark & \textbf{23.73} & \textbf{27.49} & \textbf{27.04} \\
    \bottomrule
\end{tabular}
\end{sc}
\end{small}
\end{center}
\vskip -0.1in
\label{table:shapenet}
\vspace{-0.4cm}
\end{table*}

\paragraph{2D CT reconstruction} 
For this task, we use an attention-based encoder in our PONP framework (see Section \ref{ablations}). We let the target inputs $\mathbf{x}_\mathcal{T}$ be the coordinates of the 2D image, and the target outputs $\mathbf{y}_\mathcal{T}$ be the corresponding volume densities. We also set the context output be the volume density, and we choose the context inputs to be the angles used to produce the projections, since the image coordinates do not add any extra information. We also drop the random Fourier features used by the neural field in \cite{tancik2021learned}, as we find that the performance of our method is much lower with Fourier features. For this task, we do not compare against the Transformer INR baseline \cite{chen2022transformers}, which is unsuitable to handling the 1D partial information. 

Quantitative results can be found in Table \ref{table:ct-comparison} and qualitative results can be found in Figure \ref{fig:ct}. We observe that our PONP method significantly outperforms Reptile. Visually, we also observe that the predicted CT reconstructions are much sharper with our PONP method than with Reptile. We also observe that even without test-time optimization, our neural process method is able to outperform random initialization (i.e., training a neural field from scratch) and is able to outperform Reptile in the 1, 2, and 4 view settings. This shows that with similar amounts of test-time optimization, neural processes greatly outperform gradient-based meta-learning, and in the test-time optimization free setting neural processes are still competitive, especially in setting with low amounts of partial information, while requiring zero test-time computation.

\begin{figure}[t]
\begin{center}
   \begin{tabular}{ccc}
   \includegraphics[width=0.27\linewidth]{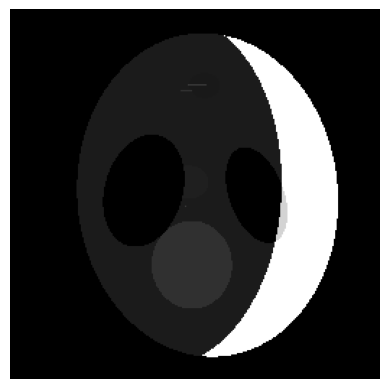} & \includegraphics[width=0.27\linewidth]{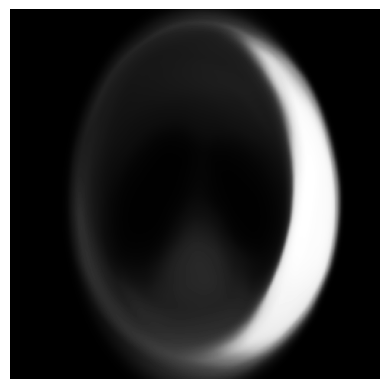} & \includegraphics[width=0.27\linewidth]{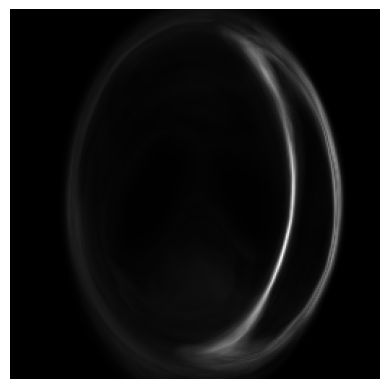} \\
   \includegraphics[width=0.27\linewidth]{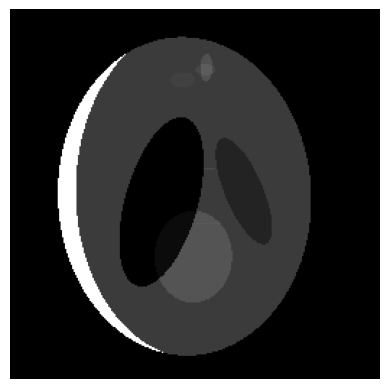} & \includegraphics[width=0.27\linewidth]{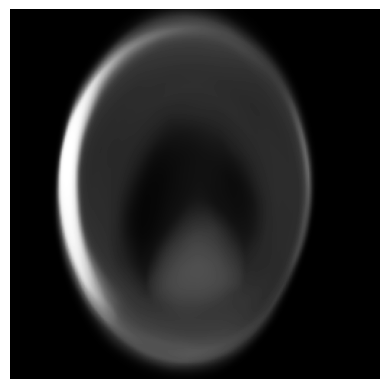} & \includegraphics[width=0.27\linewidth]{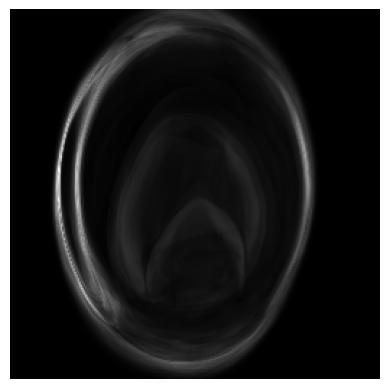} \\
   GT & Mean & Var. \\
   \end{tabular}
\end{center}
\caption{Visualizing uncertainty prediction by PONP.}
\vspace{-.5cm}
\label{fig:uncertainty}
\end{figure}

Due to probabilistic nature of our method, we can use the standard deviations learned post-forward map as a measure of uncertainty. The mean and standard deviation of reconstructions from 100 latent $\mathbf{z}$ samples from our AttnLNP CT model with 1 input view. Our model shows uncertainty about the boundaries of the different regions of the CT scan. See Figure~\ref{fig:uncertainty}.

\begin{table*}[t!]
\caption{Comparison of different neural process architectures for our PONP method on the CT reconstruction task. In each of the second and third sections of the table (corresponding to methods not using test-time optimization and methods using test-time optimization, respectively), the best performing PSNRs are bolded and the second-best PSNRs are underlined. The TTO column is checked if test-time optimization was used.}
\vskip 0.15in
\begin{center}
\begin{small}
\begin{sc}
\begin{tabular}{lccccc}
    \toprule
    Architecture & TTO & 1 Views & 2 Views & 4 Views & 8 Views \\
    \midrule
    Random Init. \cite{tancik2021learned} & \cmark & 13.63 & 14.15 & 16.31 & 21.49 \\
    \midrule
    CNP \cite{garnelo2018conditional} &  \xmark & \textbf{16.39} & \textbf{20.39} & 22.14 & 22.60 \\
    LNP \cite{garnelo2018neural} &  \xmark & \underline{16.28} & 19.43 & 21.02 & 21.34 \\ 
    AttnCNP \cite{dubois2020npf} &  \xmark & 16.24 & 20.12 & \underline{22.60} & \underline{23.60} \\
    AttnLNP \cite{kim2019attentive} &  \xmark & 16.26 & \underline{20.27} & \textbf{23.07} & \textbf{23.88} \\
    \midrule
    Reptile \cite{tancik2021learned} &  \cmark & 15.09 & 18.70 & 22.00 & 27.34 \\
    CNP \cite{garnelo2018conditional} & \cmark & \textbf{24.67} & \textbf{29.11} & \underline{37.54} & \underline{37.66} \\
    LNP \cite{garnelo2018neural} & \cmark & \underline{24.66} & \underline{26.92} & 29.22 & 29.34 \\
    AttnCNP \cite{dubois2020npf} & \cmark & 18.32 & 25.14 & \textbf{38.18} & \textbf{38.24}\\
    AttnLNP \cite{kim2019attentive} & \cmark & 18.10 & 23.42 & 32.47 & 32.94 \\
    \bottomrule
\end{tabular}
\end{sc}
\end{small}
\end{center}
\vskip -0.1in
\label{table:arch-ablation}
\vspace{-0.4cm}
\end{table*}

\paragraph{View synthesis for ShapeNet objects} 
For this task, we also report the Matched and Shuffled baselines of \cite{tancik2021learned}. Matched is an initialization that is trained from scratch to match the output of the meta-learned initialization, and Shuffled is an initialization that permutes the weights in the meta-learned initialization \cite{tancik2021learned, chen2022transformers}. Since a key part of instantiating our PONP framework for a given task is finding a suitable encoder, we adapt the vision transformer \cite{dosovitskiy2020image} encoder of the Transformer INR method. This makes sense since the vision transformer encoder can be thought of as an attention-based neural process encoder that takes is designed for the 2D image sensor observations and conditions a neural field via hypernetwork. 
To fully incorporate the vision transformer encoder into our framework, we modify the neural field to produce the mean and variance of a Gaussian distribution and apply the CNP loss (see Sec. \ref{np-framework}).

Quantitative results can be found in Table \ref{table:shapenet} and qualitative results can be found in Figure \ref{fig:shapenet}. Noting that PSNR is a on a log scale, we find that our method is comparable to or slightly outperforms the previous state-of-the-art Transformer INR baseline, while being more widely applicable. Our method also slightly improves over the Transformer INR method when the number of views increases to 2. We also find that even with only 1 view, our method clearly outperforms gradient-based meta-learning methods, even without test-time optimization.

\subsection{Ablations}
\label{ablations}

\paragraph{Architecture} In the absence of a specially-designed encoder, our PONP framework can leverage existing neural process architectures. We examine the performance of different neural process architectures on the CT reconstruction task. We compare among the CNP \cite{garnelo2018conditional}, LNP \cite{garnelo2018neural}, AttnCNP \cite{dubois2020npf}, and AttnLNP \cite{kim2019attentive} architectures. Quantitative results on the CT reconstruction experiment can be found in Table \ref{table:arch-ablation}. We find that the more recent attention-based architectures tend to outperform older MLP-based architectures when no test-time optimization is allowed. With test-time optimization, all neural process architectures outperform the Reptile and random initialization baselines. Unexpectedly, CNP architectures outperforms LNP architectures by a wide margin, which is unexpected as both the pre-test-time optimization results are similar and and LNP models do not achieve significantly better likelihoods during training. We hypothesize that this is due to sampling from the latent distribution required by LNP architectures.
Without test-time optimization, we find that all neural process architectures outperform the Reptile baseline when there are a low amount of projections and test-time optimization steps, with the best model (AttnLNP) outperforming all baselines in the 1, 2, and 4 view settings.


\section{Conclusion}

In summary, we introduce a new framework for neural field generalization that is inspired by neural processes. We show that our framework outperforms both gradient-based meta-learning and hypernetwork approaches for a variety of different neural field problems. With this work, we demonstrate the promise of neural process-based algorithms for efficiently learning neural fields. In this direction, avenues for future investigation include quantifying the predictive uncertainty learned by neural process algorithms and applications of our method to applications such as biomedical imaging and reconstruction. 

{\small
\bibliographystyle{ieee_fullname}
\bibliography{egbib}
}

\clearpage

\appendix
\section{Experimental details}
\label{experiment-deets}

All experiments were implemented in PyTorch. Our PONP framework was implemented with the help of the Neural Process Family library \citep{dubois2020npf}. All experiments were run on an NVIDIA Titan RTX or NVIDIA 2080 Ti GPU. 

\subsection{2D image regression and completion}

\begin{figure*}[htb]
\begin{center}
    \begin{tabular}{cccccc|c}
    \rotatebox{90}{Regression} & \includegraphics[width=0.1\textwidth]{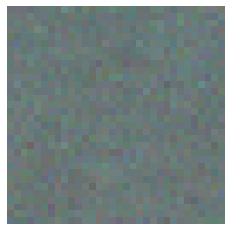} & \includegraphics[width=0.1\textwidth]{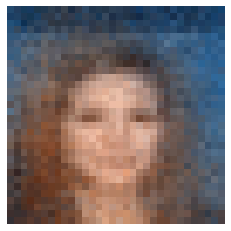} & \includegraphics[width=0.1\textwidth]{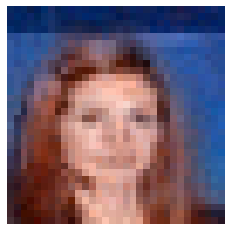} & \includegraphics[width=0.1\textwidth]{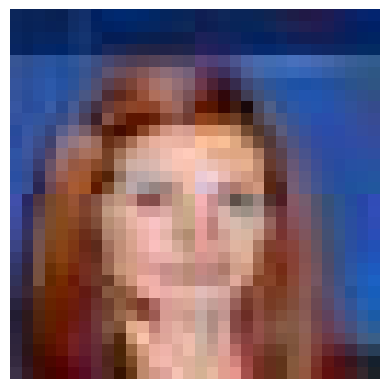} & \includegraphics[width=0.1\textwidth]{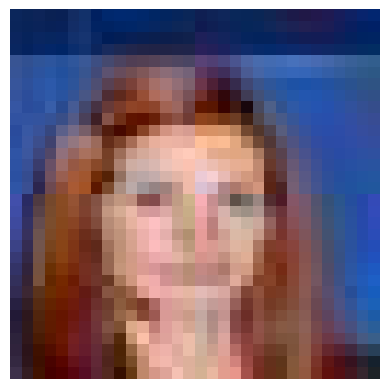} & \includegraphics[width=0.1\textwidth]{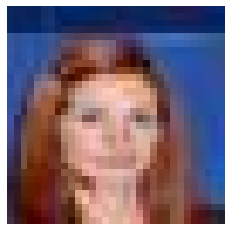} \\
    \rotatebox{90}{Completion} & \includegraphics[width=0.1\textwidth]{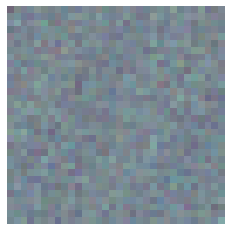} & \includegraphics[width=0.1\textwidth]{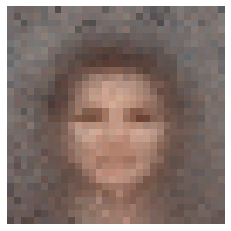} & \includegraphics[width=0.1\textwidth]{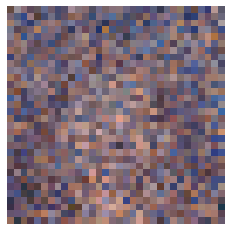} & \includegraphics[width=0.1\textwidth]{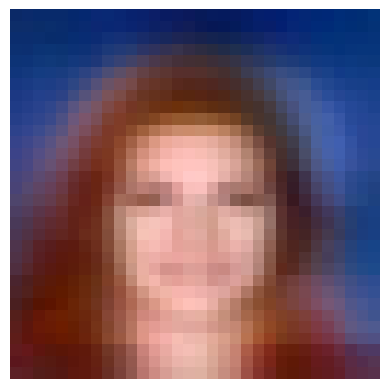} & \includegraphics[width=0.1\textwidth]{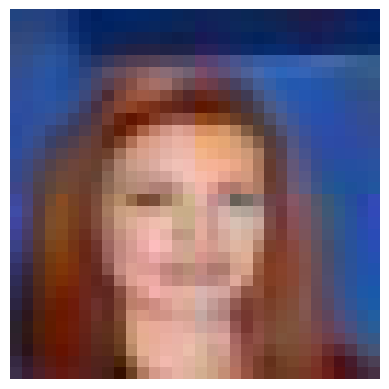} & \includegraphics[width=0.1\textwidth]{images/img_completion_ex.png}\\
    & Rand. Init. & Reptile \cite{tancik2021learned} & MAML \cite{tancik2021learned} & Trans. INR \cite{chen2022transformers} & NP (Ours) & Target \\
\end{tabular}
\end{center}
\caption{Qualitative examples of the 2D image regression and completion tasks for various methods. While the visual results for the image regression tasks are similar for meta-learning, Transformer INR, and neural processes, neural processes achieve much more accurate results than either competing method in the partially-observed image completion setting.}
\label{fig:2D-img}
\end{figure*}

The Celeb-A dataset was resized to $32 \times 32$
using Pillow's \texttt{resize} function. 

The gradient-based meta-learning baselines were re-implemented in PyTorch based on the original public JAX implementations. Both baselines used a SIREN \cite{sitzmann2019scene} architecture was used with $\omega_0 = 200$ and five hidden layers of 128 channels, and used the same initialization as the SIREN paper. The MAML baseline was trained for 200K iterations with an outer batch size of 3, an outer loop learning rate of $10^{-5}$, and an inner loop learning rate of $10^{-2}$. The outer loop used the Adam optimizer, while the inner loop used SGD and two gradient steps. The Reptile baseline was trained for 160K steps and the same learning rates as MAML but with an outer batch size of 10. The parameter choices for both gradient-based optimization methods were based on \cite{tancik2021learned}, with the exception of the number of steps. Similarly, as in \cite{tancik2021learned} the random initialization baseline was trained with the Adam optimizer, a learning rate of $10^{-2}$, and 2 gradient steps.

The Transformer INR baseline \cite{chen2022transformers} was run using the official public code repository with 128 channels and 128 channels in the MLP, with a patch size of 3 in the input tokenizer. All other parameters are the same as the original implementation. 

Our PONP used a ReLU MLP neural field with 128 channels and was trained with the Adam optimizer with a learning rate of $10^{-3}$ with exponential learning rate decay to a final learning rate of $10^{-4}$ and a batch size of 32.

\subsection{CT reconstruction}

Our PONP neural field was identical to the one used in \cite{tancik2021learned} except we did not use random Fourier features, as we found that the results were much better without the random Fourier features. Our method was trained with a learning rate of $10^{-4}$ with exponential learning rate decay to a final learning rate of $10^{-5}$ and a batch size of 1. Test time optimization was done with the Adam optimizer and a learning rate of $10^{-4}$. As in \cite{tancik2021learned}, in the 1, 2, 4, and 8 view cases all methods used 50, 100, 1000, and 1000 test-time optimization steps, respectively. 

The baseline results were taken from \cite{tancik2021learned}.

\subsection{ShapeNet view synthesis}

Our PONP model was trained with learning rate $10^{-4}$, batch size 32, and 1000 epochs. As in \cite{chen2022transformers}, we only use 100 test-time optimization steps. The gradient-based meta-learning baselines \cite{tancik2021learned} used 1000 or 2000 test-time optimization steps depending on the number of training views and the category of object. The exact number of test-time optimization steps for each baseline can be found in Appendix A of \cite{tancik2021learned}.

All baselines results were taken from their respective papers (\cite{tancik2021learned} and \cite{chen2022transformers}).

\section{More ablations of 2D image experiments}
\label{more-ablations}

\begin{table}[ht!]
\caption{Comparison of different neural process architectures for our neural process method on the 2D image regression and 2D image completion tasks. Parameters refers to the total number of parameters for the model, including the neural field itself. The best results for each different setting are bolded and the second best results are underlined.}
\vskip 0.15in
\begin{center}
\begin{small}
\begin{sc}
\begin{tabular}{lccc}
    \toprule
    Architecture & Reg. PSNR & Comp. PSNR & Params \\
    \midrule
    CNP \cite{garnelo2018conditional} & 20.37 & 18.32 & 367K \\
    LNP \cite{garnelo2018neural} & 19.55 & 19.23 & 417K \\
    AttnCNP \cite{dubois2020npf} & \underline{65.88} & 22.30 & 386K \\
    AttnLNP \cite{kim2019attentive} & 61.84 & \underline{22.68} & 468K\\
    ConvCNP \cite{gordon2019convolutional} & \textbf{78.87} & \textbf{23.24} & 340K \\
    \bottomrule
\end{tabular}
\end{sc}
\end{small}
\end{center}
\vskip -0.1in
\label{table:img-comp-ablation}
\end{table}

We compare the CNP \cite{garnelo2018conditional}, LNP \cite{garnelo2018neural}, 
AttnCNP \cite{dubois2020npf}, AttnLNP \cite{kim2019attentive}, and ConvCNP \cite{gordon2019convolutional} architectures for neural processes. The LNP and AttnLNP models were trained with NPML \cite{foong2020meta}. 

Quantitative results on the 2D image regression and 2D image completion tasks can be found in Table \ref{table:img-comp-ablation}. For the image regression task, we find that the earliest architectures (CNP, LNP) fail to out-perform the gradient-based meta-learning and Transformer INR baselines on the 2D image regression task, but that the more advanced neural process architectures (AttnCNP, AttnLNP, ConvCNP) greatly out-perform all baselines. This is consistent with previous literature, which found that CNPs were unable to out-perform gradient-based meta-learning for learning simple signed-distance functions \cite{sitzmann2020metasdf}. On the image completion task, we find that all neural process architectures outperform gradient-based meta-learning methods, and all neural process architectures outperform the Transformer INR hypernetwork method with the exception of the CNP architecture. This shows that the superiority of neural processes is not just due to architectural choices.

\end{document}